\documentclass[12pt]{article}
 \pdfoutput=1

\usepackage[sort]{natbib}
\bibpunct[, ]{(}{)}{;}{a}{,}{,}

\usepackage{amsmath}
\usepackage{physics}
\usepackage{float}
\newcommand{\blind}{0}

\usepackage{url}            
\usepackage{booktabs, makecell, multirow, tabularx}       
\usepackage{mathtools}  
\usepackage{nameref, hyperref, cleveref}
\usepackage{bbm}
\usepackage{bm}
\usepackage{amsthm}
\theoremstyle{plain}
\newtheorem{theorem}{Theorem}[section]

\newtheorem{corollary}[theorem]{Corollary}

\theoremstyle{definition}

\theoremstyle{remark}
\newtheorem{remark}{Remark}

\newcommand{\RR}{I\!\!R} 

\newcommand{\bigo}[1]{\mathcal{O}\left( #1 \right)}
\newcommand{\bigop}[1]{\mathcal{O}_P\left( #1 \right)}

\DeclareMathAlphabet{\mathpzc}{OT1}{pzc}{m}{it}

\usepackage{xr}
\makeatletter
\newcommand*{\addFileDependency}[1]{
  \typeout{(#1)}
  \@addtofilelist{#1}
  \IfFileExists{#1}{}{\typeout{No file #1.}}
}
\makeatother

\newcommand*{\myexternaldocument}[1]{
    \externaldocument{#1}
    \addFileDependency{#1.tex}
    \addFileDependency{#1.aux}
}
\myexternaldocument{Appendix}

\begin{document}

\def\spacingset#1{\renewcommand{\baselinestretch}%
{#1}\small\normalsize} \spacingset{1}
\newcommand{\scriptsj}[1]{\mathpzc{s}_j}


\if0\blind
{
  \title{\bf Approximate cross-validated mean estimates for Bayesian hierarchical regression models}
  \author{Amy Zhang\\
    Department of Statistics, Pennsylvania State University\\
    and \\
    Michael J. Daniels \\
    Department of Statistics, University of Florida \\
    and\\
    Changcheng Li\\
    School of Mathematical Sciences, Dalian University of Technology\\
    and\\
    Le Bao \\
    Department of Statistics, Pennsylvania State University
    }
  \maketitle
} \fi

\if1\blind
{
  \bigskip
  \bigskip
  \bigskip
  \begin{center}
    {\LARGE\bf Approximate Cross-validated Mean Estimates for Bayesian Hierarchical Regression Models}
\end{center}
  \medskip
} \fi

\bigskip
\begin{abstract}
We introduce a novel procedure for obtaining cross-validated predictive estimates for Bayesian hierarchical regression models (BHRMs). BHRMs are popular for modeling complex dependence structures (e.g., Gaussian processes and Gaussian Markov random fields) but can be computationally expensive to run. Cross-validation (CV) is, therefore, not a common practice to evaluate the predictive performance of BHRMs. Our method circumvents the need to re-run computationally costly estimation methods for each cross-validation fold and makes CV more feasible for large BHRMs. We shift the CV problem from probability-based sampling to a familiar and straightforward optimization problem by conditioning on the variance-covariance parameters. Our approximation applies to leave-one-out CV and leave-one-cluster-out CV, the latter of which is more appropriate for models with complex dependencies.
In many cases, this produces estimates equivalent to full CV. We provide theoretical results, demonstrate the efficacy of our method on publicly available data and in simulations, and compare the model performance with several competing methods for CV approximation. Code and other supplementary materials available online.
\end{abstract}

\noindent%
\vfill

\newpage
\spacingset{1.5} 

\section{Introduction}

Bayesian hierarchical regression models (BHRMs) are often used to model complex dependence structures while producing probabilistic uncertainty estimates. Except for the simplest models, BHRMs require computationally expensive methods such as Markov Chain Monte Carlo (MCMC) to obtain the posterior density. It has led to many papers that either present new methods to approximate the posterior density \cite[e.g.,][]{kingma2013auto, lewis1997estimating, rue2009approximate} or attempt to make current methods more efficient \citep{bardenet2017markov, korattikara2014austerity, quiroz2019speeding}. 

The computational cost of BHRM increases by an order of magnitude when we need to repeat the posterior density estimation process for each fold of the cross-validation (CV). {A common approach for cross-validation is to randomly partition the data into $K$ equally-sized subsets, called $K$-fold CV. Reducing the number of CV folds, $K$, can mitigate the cost of cross-validation.} However, when the data are not independent of each other, random $K$-fold cross-validation can select models which overfit because of the high correlation between test and training data \citep{arlot2010survey, opsomer2001nonparametric}. {Models with repeated measures are one such example, where the correlation between repeated samples from the same unit means that random $K$-fold cross-validation can select models which overfit the unit-specific effects. This produces over-optimistic estimates of the selected model's performance on a new unit. CV dropping out whole clusters (possibly more than one) at a time would be the most robust, but it often requires far more than 10 CV folds.}

{We reduce the computational cost of BHRM in the CV setting by introducing a novel procedure for obtaining BHRM posterior mean estimates. We use the full-data posterior mean of the variance-covariance parameters as a plug-in estimator and re-estimate all other model parameters given the training data and the plug-in estimate. Our method is motivated by \citet{kass1989approximate} who approximated the posterior mean $E(X \beta | Y)$ using the conditional mean given the estimated variance terms. Our extensions include (1) approximating hyperparameters in the CV setting, (2) allowing the model to have random effects with any dependence structure, (3) extending the approximation from Bayesian hierarchical linear regression models to generalized linear mixed models, and (4) considering the leave-one-cluster-out CV which is more appropriate for models with complex dependency. We show with multiple datasets that our method produces more stable estimates than other competing methods while keeping computation time reasonable.}

In Section~\ref{sec:axe}, we introduce our procedure, which we refer to as AXE, an abbreviation for (A)pproximate (X)cross-validation (E)stimates. It can be applied to both linear mixed models and generalized linear mixed models, and to many CV schemas, e.g., $K$-fold, leave-one-out (LOO), and leave-one-cluster-out (LCO). We also discuss AXE's asymptotic properties and finite sample behavior. In Section~\ref{sec:other_lco}, we summarize competing methods for CV approximation and extend some so that they can be applicable to LCO-CV. Section~\ref{sec:reslts_summary} presents and discusses empirical results comparing the manual cross-validation (MCV) and different approximate CV estimates using a variety of publicly available data sets. In Section~\ref{sec:discussion}, we summarize the method and our findings.
 
\section{Approximate cross-validation estimates using plug-in estimators (AXE)}\label{sec:axe}

Section~\ref{sec:lmm} presents the AXE procedure in the linear regression setting. Section~\ref{sec:glmm} presents the AXE procedure in the generalized linear regression setting, which requires an additional approximation.
A particular challenge in the LCO-CV approximation is that all observations informing the estimation of the random effect are associated with the test data which we describe in detail in Section~\ref{sec:lco}. Section~\ref{sec:asymptotic} describes the large-sample behavior of AXE estimates. Section~\ref{sec:finite} provides a finite-sample diagnostic, which we call AXE+.

We introduce the notation used throughout the paper. Let $Y \in \RR^N$ denote a continuous response vector and $X \in \RR^{N\times P}$ denote the design matrix. For the $J$-fold cross-validation, $j = 1, \dots, J$ indicates the cross-validation folds; $\mathpzc{s}_j$ corresponds to the indices of the held-out data for CV fold $j$; $n_j$ is the corresponding sample size; $Y_{\mathpzc{s}_j}\in \RR^{n_j}$ is the test response data; $Y_{-\mathpzc{s}_j}\in \RR^{N-n_j}$ the training response data; and similarly $X_{\mathpzc{s}_j} \in \RR^{n_j \times P}$ refers to the rows of $X$ indexed by $\mathpzc{s}_j$, while $X_{-\mathpzc{s}_j} \in \RR^{(N-n_j) \times P}$ refers to the remaining rows of $X$ without $X_{\mathpzc{s}_j}$. We denote the transpose of a matrix (or vector) $A$ as $A'$.
 
\subsection{AXE for linear mixed models}\label{sec:lmm}

In linear mixed models (LMM), the continuous response vector, $Y$, follows
\begin{equation}
 Y | \beta, \phi \sim N(X\beta, \phi^2),
\label{eqn:normlinreg}
\end{equation}
where $X$ are the non-random covariates, $\beta \coloneqq (\beta_1', \beta_2')' \in \RR^{P}$ includes fixed effects $\beta_1$ and random effects $\beta_2$, $\phi \in \RR^{+}$ is a positive scalar. 
\begin{equation}
\begin{split}
 \beta \sim N\left(\alpha, \begin{bmatrix}C & 0 \\ 0 & \Sigma\end{bmatrix}\right),\\
 \Sigma  \sim f_{\Sigma}(\Sigma), \quad \phi \sim f_{\phi}(\phi),
\end{split}
\label{eqn:hyper}
\end{equation}
where $C \in \RR^{P_1 \times P_1}$ is a fixed positive-definite matrix and often a diagonal matrix with large positive values. The hyperparameters, $\Sigma \in \RR^{P_2 \times P_2}$, can take many forms, as long as it is positive-definite, e.g., models with Gaussian Markov random fields sample from the space of all possible $\Sigma$'s such that a variable in $\beta_2$ is independent of all others, given its neighborhood. 
{Note that $f_{\phi}(\phi)$ in our setting is a fixed proper prior. Hence $\phi = \bigop{1}$, which means that for any $\epsilon>0$, there exists $\delta > 0$ such that $P(|\phi|\leq \delta) > 1-\epsilon$. In addition, from $\phi>0$, we know that $\phi^{-1} = \bigop{1}$. The probabilities for $\phi$ and $\phi^{-1}$ come from their prior distributions following the Bayesian point of view. }
Finally, we take $\alpha = 0$ throughout this paper without loss of generality.

The posterior mean for $\beta$ conditioned on variance parameters has the form
\begin{equation}
\label{eqn:axe_linregr}
   E(\beta | \Sigma, \phi, Y) =  \phi^{-2}VX'Y,  \quad V = \left(\phi^{-2}X'X + \begin{bmatrix}C^{-1} & 0 \\ 0 & \Sigma^{-1}\end{bmatrix}\right)^{-1},
\end{equation}
which is analogous to the form of a Frequentist linear regression. We propose to use the full-data posterior mean estimates $\hat{\Sigma} = E(\Sigma | Y)$ and $\hat{\phi} = E(\phi | Y)$ as plug-in estimators in (\ref{eqn:axe_linregr}) to derive cross-validated mean estimates. In the linear regression setting, this results in a simple and straightforward closed-form solution for the cross-validated mean estimates:
\begin{equation}\label{eqn:axe}
    \hat{Y}_{\mathpzc{s}_j}^{\text{AXE}} = E(X_{\mathpzc{s}_j}\beta | Y_{-\mathpzc{s}_j}, \hat{\Sigma}, \hat{\phi}) = \hat{\phi}^{-2}X_{\mathpzc{s}_j}\left(\hat{\phi}^{-2}X_{-\mathpzc{s}_j}'X_{-\mathpzc{s}_j} + \begin{bmatrix}C^{-1} & 0 \\ 0 & \hat{\Sigma}^{-1}\end{bmatrix}\right)^{-1}X_{-\mathpzc{s}_j}'Y_{-\mathpzc{s}_j}.
\end{equation}
The posterior variance of fixed effects $\beta_1$, $C$, is not estimated and is taken as the diagonal matrix of prior variances for $\beta_1$. One common practice is to take $C^{-1}$ as the matrix of $0$s, which corresponds to the assumption that the fixed effects have infinite variance. 

{The AXE procedure in (\ref{eqn:axe}) essentially proposes using a Frequentist cross-validation to estimate $E(Y_{\mathpzc{s}_j} | Y_{-\mathpzc{s}_j})$ in place of a full Bayesian estimation of the posterior predictive density. AXE is likewise $\bigo{N^2P + P^3}$ in time for each CV fold. In comparison, Gibbs sampling of the same problem (when available) is $\bigo{MN^3P + MNP^2 + MP^3}$ in time for each CV fold, where $M$ is the number of MCMC iterations.}

Our approach relies on the approximation
that the conditional expectation of the posterior mean evaluated at the full-data posterior means of the variance parameters,  $\hat{\Sigma}= E(\Sigma | Y)$ and  $\hat{\phi} = E(\phi | Y)$,
approximates the desired posterior predictive mean, i.e., $E(X_{\mathpzc{s}_j}\beta | Y_{-\mathpzc{s}_j}, \hat{\Sigma}, \hat{\phi}) \approx E(X_{\mathpzc{s}_j}\beta | Y_{-\mathpzc{s}_j})$.  This looks similar to the approximation in \citet{kass1989approximate} but here we are in a cross-validation setting and we are plugging in the full-data estimates (as opposed to the training data estimates).  
In Section~\ref{sec:asymptotic}, we show that the  approximation holds 
asymptotically under leave-one-cluster-out cross-validation.

\subsection{{AXE for generalized linear mixed models}}\label{sec:glmm}

Generalized linear mixed models (GLMMs) assume that $Y | \beta$ is not normally distributed. We follow the approximate inference in GLMM developed by \cite{breslow1993approximate} and derive the corresponding AXE for GLMM. Suppose $Y|\beta$ are independent with some probability density function $\pi$, so that 
\begin{equation} 
\label{eqn:glmreg}
    Y_i | X_i \beta \sim \pi(u_i, a_i, \phi^2, v(u_i)),
\end{equation}
where $E(Y_i | X_i \beta) = u_i$, $\mathrm{var}(Y_i | X_i \beta) = a_i \phi^2 v(u_i)$, $a_i$ is a known constant, $\phi^2$ is a dispersion parameter. The regression model assumes that $g(u_i) = X_i \beta$, where $g(.)$ is the link function and the prior distribution of $\beta$ is specified in (\ref{eqn:hyper}).

Note that, the normal priors on the coefficients $\beta$ are no longer conjugate, and the analytic solution of (\ref{eqn:axe}) is unavailable. Instead, an iterative weighted least square algorithm (IWLS) is commonly used to fit GLM or GLMM \citep{holland1977robust,green1984iteratively}. Defining a working response,
\begin{equation*} 
   \tilde{Y_i} =  X_i\beta + (Y_i-u_i) g'(u_i),
\end{equation*}
and a $n \times n$ diagonal weight matrix, $W$, with diagonal terms
\begin{equation*} 
   w_i = \{ \phi^2 a_i v(u_i) [ g'(u_i) ]^2 \} ^{-1},
\end{equation*}
i.e., for the logistic model, $g(u_i) = \log(u_i) - \log(1-u_i)$, $g'(u_i)=\frac{1}{u_i(1-u_i)}$, and $w_i = u_i(1-u_i)$; for the Poisson model, $g(u_i) = \log(u_i)$, $g'(u_i)=\frac{1}{u_i}$, and $w_i = u_i$. IWLS repeatedly updates the working response vector and the weight matrix until its convergence. Given $W$, \cite{harville1977maximum} shows that the best linear unbiased estimation (BLUE) of $\beta$ can be obtained from the associated weighted linear model, 
\begin{equation*} 
   \tilde{Y} | \beta, \sigma^2 \sim N(X\beta, W^{-1}).
\end{equation*}

In the cross-validation setting, we approximate the training data weights, $W_{-\mathpzc{s}_j}$, by the corresponding weights at the convergence of the full data analysis. It is reasonable under the conditions we present in Section \ref{sec:lmm}: 1) $E(X_{\mathpzc{s}_j}\beta | \tilde{Y}_{-\mathpzc{s}_j}, \dot{\Sigma}, \dot{\phi}) \approx E(X_{\mathpzc{s}_j}\beta | \tilde{Y}_{-\mathpzc{s}_j})$, where $\dot{\Sigma} = E(\Sigma | \tilde{Y}_{-\mathpzc{s}_j})$, $\dot{\phi} = E(\phi | \tilde{Y}_{-\mathpzc{s}_j})$ and 2) $E(\Sigma | \tilde{Y}) \approx \dot{\Sigma}$, $E(\phi | \tilde{Y}) \approx \dot{\phi}$.

We plug in the working response vector, $\tilde{Y}_{-\mathpzc{s}_j}$, the weight matrix, $W_{-\mathpzc{s}_j}$, and other variance estimates, $\hat{\Sigma}$, in the full data analysis, and obtain the following AXE estimates for the test data working response, $\tilde{Y}_{\mathpzc{s}_j}$,
\begin{equation}
     \hat{\tilde{Y}}_{\mathpzc{s}_j}^{\text{AXE}} = E(X_{\mathpzc{s}_j}\beta | \tilde{Y}_{-\mathpzc{s}_j}, W_{-\mathpzc{s}_j}, \hat{\Sigma}) = X_{\mathpzc{s}_j}\left(X_{-\mathpzc{s}_j}'W_{-\mathpzc{s}_j}X_{-\mathpzc{s}_j} + \begin{bmatrix}C^{-1} & 0 \\ 0 & \hat{\Sigma}^{-1}\end{bmatrix}\right)^{-1}X_{-\mathpzc{s}_j}'W_{-\mathpzc{s}_j}\tilde{Y}_{-\mathpzc{s}_j}.
\end{equation}
Finally, we transform the test data working response to the original scale by $\hat{{Y}}_{\mathpzc{s}_j}^{\text{AXE}} = g^{-1}( \hat{\tilde{Y}}_{\mathpzc{s}_j}^{\text{AXE}})$. If there is any concern that the weight matrix might differ between the full-data analysis and the training data analysis, then we can run a few more iterations of the IWLS algorithm. We start from the AXE estimates of the working responses and iteratively update the training data weights and the working responses. In practice, $\hat{\tilde{Y}}_{\mathpzc{s}_j}^{\text{AXE}}$ without additional IWLS iterations works well.

\subsection{Leave-one-cluster-out cross-validation (LCO-CV)}\label{sec:lco}
One can apply AXE to any CV design, of which leave-one-out (LOO-CV) and leave-one-cluster-out (LCO-CV) are popular choices. For both LOO-CV and LCO-CV, the number of CV folds can grow with the amount of data, making the computation expensive. Thus they typically benefit the most from CV approximation methods. We focus on proving AXE convergence under LCO-CV, of which LOO-CV is a special case. We first define LCO-CV and related notation. Then, we introduce and discuss existing LCO-CV approximation methods.

In LOO-CV, a single observation is held out as test data for each CV fold, and the number of CV folds equals the number of data points, $N$. LOO-CV is commonly used because it maintains as much similarity to the full data model posterior as possible while evaluating the fit to new data. LCO-CV is often used in models with clustered data and cluster-specific parameters, e.g., the patient-specific random effects in a longitudinal model with $J$ patients. When the data are sampled uniformly at random to form CV folds, data from one patient can be split between the training and test data. It introduces correlation between the training and test data and can result in overfitting to the patient-specific parameters \citep{arlot2010survey, opsomer2001nonparametric}. To fairly evaluate the predictive capability for a new cluster, LCO-CV defines $J$ CV folds corresponding to $J$ clusters and withholds all observations informing the estimation of a cluster-specific random effect.

LCO-CV has been a challenging CV design for approximate inference. Existing CV approximation methods use the full-data posterior density $f( \beta | Y)$ to approximate the training data posterior density $f(\beta | Y_{-\mathpzc{s}_j})$. Under LCO-CV, these approximations can be inaccurate for two reasons. First, since the test data corresponds to the data in a cluster, the test data sample size can be large. It is shown that the larger the sample size of the test data, the more likely it is that the difference in densities $f(\beta | Y)$ and $f(\beta | Y_{-\mathpzc{s}_j})$ is large  \citep{gelfand1992model}. Second, under cross-validation, the cluster-specific random effects, $\theta_j$, cannot be estimated from the training data, relying instead on samples drawn from $ \theta_j | \theta_{-j}, \Sigma$.

To focus on the random effects which pertain to the LCO-CV design, we split $\beta$ into $\theta$ and the remaining coefficients in $\beta$: 
\begin{equation}
\label{eqn:def_mutheta}
    \beta = (\beta_{/\theta}', \hspace{0.5em} \theta')',
\end{equation}
where $\theta$ refers to the part of random effects that pertain to the CV design, and $\beta_{/\theta}$ refers to the components in $\beta$ other than $\theta$, which contain all fixed effects and the part of random effects that do not pertain to the CV design. $\theta \coloneqq (\theta_1, \theta_2,  \ldots, \theta_J)' \in \RR^{K}$ ($J \le K \le P_2$). Similarly. we split $X$ column-wise into
\begin{equation}
    \label{eqn:def_Xmutheta}
    X = [X_{\beta_{/\theta}} \hspace{0.5em} X_{\theta}],
\end{equation}
where $X_{\theta}$ contains the subset of covariates that correspond to $\theta$ and $X_{\beta_{/\theta}}$ contains the remaining columns of $X$.

\subsection{Asymptotic behavior under LCO-CV}
\label{sec:asymptotic}

For the linear mixed models (LMM) case, we first show that the full-data posterior means for variance parameters $\Sigma$ and $\phi$ approximate the training data posterior means; see Theorem \ref{thm:varconv}. We then show the convergence of the AXE approximation in Corollary \ref{cor:axe}.

\begin{theorem}
\label{thm:varconv}
Let response vector $Y \in \RR^N$ of a hierarchical linear regression follow a normal distribution as in (\ref{eqn:normlinreg}) and (\ref{eqn:hyper}) and define $\theta$ as in (\ref{eqn:def_mutheta}). The data are partitioned into $J$ CV folds based on $\theta$, where all data informing $\theta_j$ correspond to the test data for CV fold $j$, and $\mathpzc{s}_j \subset \{1, \dots, N\}$ is the set of indices for the test data in the $j^{th}$ CV fold. Let $X_{\mathpzc{s}_j} = \mathbbm{1}_{n_j} x_j'$ for some vector $x_j \in \RR^P$, where $\mathbbm{1}_{n_j}$ is a vector of $1$s with length $n_j$, $n_j$ is the size of $\mathpzc{s}_j$, and $n_j\geq 1$, $\mathbbm{1} \in \text{span}(X)$, and $f_{\Sigma}(\Sigma)$ and $f_{\phi}(\phi)$ are proper prior densities. $V$ is defined as in (\ref{eqn:axe_linregr}). {As $J$ goes to infinity (and thus $N$ goes to infinity), we have
\begin{eqnarray*}
|E(\phi | Y_{-\mathpzc{s}_j})^{-1} E(\phi | Y) -1| &=&  \bigop{n_j/N},\\
\|E(\Sigma | Y_{-\mathpzc{s}_j})^{-1} E(\Sigma | Y) - I\| &=& \bigop{n_j/N},
\end{eqnarray*}
where $I$ is the identity matrix, $\|.\|$ is the operator norm of a matrix defined as $\|A\|=\sqrt{\lambda_1(A' A)}$ and $\lambda_1$ is the largest eigenvalue.}
\end{theorem}

For proof, see Appendix A, included in Supplemental Materials.
\begin{remark} 
{Throughout this article, the expectations, such as $E(. | Y_{-\mathpzc{s}_j})$ and $E(. | Y)$, are functions of $Y$ which are random variables. Thus, $\mathcal{O}_P{}$ refers to the randomness in $Y$ and the data generation process has been defined in  (\ref{eqn:normlinreg}) and (\ref{eqn:hyper}).}
\end{remark}

\begin{remark}
{The requirement that $X_{\mathpzc{s}_j} = \mathbbm{1}_{n_j} x_j'$ is a technical one so that an analytical form of the difference in $\ell(\Sigma, \phi | Y_{-\mathpzc{s}_j}) - \ell(\Sigma, \phi | Y)$ can be used to establish the above asymptotic properties, where $\ell(\Sigma, \phi | Y_{-\mathpzc{s}_j})$ and $\ell(\sigma,\phi |Y)$ are the log posterior densities of $\Sigma$ and $\phi$ given $Y_{-\mathpzc{s}_j}$ and $Y$ respectively. 
This condition holds in many scenarios such as repeated measure design.
We also believe this condition to be stronger than required in actual practice and include multiple real data examples which violate this condition, for which AXE performs well. See the ESP, SLC, and SRD examples in Section~\ref{sec:reslts_summary}.} The requirement that the resulting posterior distributions are proper is a mild condition that is easily satisfied in the case of normal priors on $\beta$.
\end{remark}

\begin{remark}
{When the clusters are of similar sizes, the rate $\bigop{n_j/N}$ simplifies to the rate $\bigop{1/J}$. However, the rate $\bigop{n_j/N}$ accounts for cases where clusters vary significantly in size, and it remains valid in unbalanced settings. This remark also applies to Corollary~\ref{cor:axe}.}
\end{remark}

\begin{remark}
{In our experimental results, we have found that large $J$ is more critical for the accuracy of AXE than having large $n_j$. In the Radon subsets example, we varied both $J$ from 3 - 12 and the percentage of test data from 30\% to 70\%. We found little difference in AXE error as the test data percentage increased, but large reductions in AXE error as $J$ increased.} This is because AXE error, in most cases, comes from a large variance in the estimate for $\Sigma$ across CV folds, and
$\Sigma$ is independent of the data given  $\beta$. Removing $n_j$ test observations only removes the information from one realization of $\theta_j$ for the estimation of $\Sigma$, out of $J$ total such realizations. 
\end{remark}
\begin{remark}
The proof for the convergence of $\phi$ is an extension of the proof for $\Sigma$ and so is presented with the same error bound. In most cases, we expect $\phi$ to have a narrower error bound than $\Sigma$, as all observations inform the estimation of $\phi$ equally and the overall sample size, $N$, is more critical for the estimation of $\phi$ than the number of clusters, $J$. We found this to be the case in our example applications; $\phi$ was generally well-estimated with a relatively narrow density for $\phi | Y$ and correspondingly smaller difference between $E(\phi | Y_{-\mathpzc{s}_j})$ and $\hat{\phi}$. {However, we do expect $n_j$ to be more critical for the convergence of $\phi$ than $J$.}

\end{remark}
\begin{remark}
We do not assume any specific form for $\Sigma$, which may consist of $P_2(P_2 + 1)/2$ separate random variables or, in many cases, $\Sigma$ may be a function of a much smaller number of parameters, as when $\Sigma = \sigma^2I$. Other examples include Gaussian processes or Gaussian Markov random fields, where $\Sigma$ is a covariance function based on fixed data properties with typically only 1-3 parameters that need to be estimated.  In the latter case, we expect a relatively modest number of clusters $J$ to be sufficient to provide good results, as AXE performs best when $\Sigma$ is well-estimated with a narrow posterior density, which occurs more often when $J >>$ the number of parameters for $\Sigma$. In the Radon subsets example in Section~\ref{sec:reslts_summary}, we found $J \ge 9$ to be large enough for good results in the case of 1 parameter for $\Sigma$.
\end{remark}
\begin{remark}
Similarly, we do not assume any specific prior for $\Sigma$ or $\phi$. One of the appeals of Bayesian modeling is its flexibility, and for regression, this often occurs through placing different priors on $\Sigma$, e.g., variable selection or Bayesian penalized splines. There is also a growing preference for half-t priors over the conjugate inverse-gamma \citep{gelman2006prior, polson2012half}. These considerations make it pragmatic to provide proofs for unspecified $f_{\phi}(\phi)$ and $f_{\Sigma}(\Sigma)$.  
\end{remark}

As the AXE estimate is a straightforward function of $\Sigma$ and $\phi$, we can use the result of Theorem~\ref{thm:varconv} above to derive an overall error bound for $E(X_{\mathpzc{s}_j}\beta | Y_{-\mathpzc{s}_j}, \hat{\Sigma}, \hat{\phi})$.

\begin{corollary}\label{cor:axe} Let $\hat{\Sigma}$ denote the full-data posterior mean, $E(\Sigma | Y)$, and $\tilde{\Sigma}$ the CV posterior mean over the training data $E(\Sigma | Y_{-\mathpzc{s}_j})$ for CV fold $j$.
{Note that $X_{\mathpzc{s}_j}$ is made up of identical rows and $x_{\mathpzc{s}_j}'$ refers to any row in $X_{\mathpzc{s}_j}$.}
Under the same conditions as Theorem \ref{thm:varconv}, $E(x_{\mathpzc{s}_j}'\beta | Y_{-\mathpzc{s}_j}) = E(x_{\mathpzc{s}_j}'\beta | Y_{-\mathpzc{s}_j}, \hat{\Sigma}, \hat{\phi})(1 + \bigop{n_j/N})$ 
as $J$ goes to infinity.
\end{corollary}

For proof, see Appendix A.

As the conditional expectation approximates the posterior predictive mean with error rate $\bigop{n_j/N}$, AXE approximates the posterior predictive mean with overall error rate $\bigop{n_j/N}$. In the LMM case, we have generally found that the accuracy of AXE largely depends on the accuracy of using the posterior means of the variance parameters plug-in estimators. 
In the GLMM case, AXE introduces the same approximations to the IWLS algorithm. 

\subsection{Finite sample performance}\label{sec:finite} 

In general, when $\Sigma$ and $\phi$ are estimated with narrow posterior densities, $E(\Sigma | Y_{-\mathpzc{s}_j})$ and $E(\phi | Y_{-\mathpzc{s}_j})$ will not deviate much from their full-data posterior estimates and AXE will be accurate. This is also when LCO-CV is the most expensive {with a large number of clusters} and AXE provides the most benefit. In our experience with AXE, even when posterior densities are wide, we have found that AXE can perform well (e.g., Radon subsets and Eight schools examples, see Figure \ref{fig:lrr_perc}). The cases where AXE is inaccurate are typically those with a low number of clusters or severe data imbalance such that removal of the test data severely impacts the estimation of $\Sigma$ or $\phi$. 

Intuitively, when the number of clusters is low, running manual cross-validation (MCV) may not be of concern. If additional assurance is needed, we suggest running MCV on a randomly selected subset of CV folds, stratified by clusters. By comparing the MCV point estimates to the AXE approximation, we can determine how similar the conclusions derived from the AXE approximation are to those using the MCV estimates. One common model evaluation criteria is the root mean square error (RMSE), where the error is the difference between $\hat{Y}_{\mathpzc{s}_j}^{(\text{MCV})} = E(Y_{\mathpzc{s}_j} | Y_{-\mathpzc{s}_j})$ and $Y_{\mathpzc{s}_j}$. We take the log ratio of AXE-approximated RMSE to ground-truth MCV RMSE,
\begin{equation}\label{eqn:lrr}
  \text{LRR}_{j} = \log \left(\frac{\sum_{i = 1}^{n_j} (\hat{Y}_{i}^{(\text{AXE})} - Y_{i})^2}{\sum_{i = 1}^{n_j} (\hat{Y}_{i}^{(\text{MCV})} - Y_{i})^2}\right),
\end{equation}
where LRR$_j$ is calculated separately for each chosen CV fold $j$ to obtain more fine-grained comparisons. {A low $|LRR|$ represents high similarity between the approximate RMSE and ground-truth MCV RMSE. We use LRR as our criterion rather than $\sum (\hat{Y}^{\text{AXE}}-\hat{Y}^{MCV})^2$, because the magnitude of the latter depends on the standard error for $\hat{Y}^{\text{AXE}}$; it is possible that the approximate RMSE is close to the true RMSE while $\sum (\hat{Y}^{\text{AXE}}-\hat{Y}^{MCV})^2$ is large.} We also use LRR to compare AXE approximations to other LCO methods in Section \ref{sec:reslts_summary} by replacing $\hat{Y}_{i}^{(\text{AXE})}$ in (\ref{eqn:lrr}) with the alternative LCO method's point estimate for $Y_{i}$.

If exchangeability is assumed between clusters, then the sample mean and variance of $\text{LRR}_{j}$'s can provide inference for the expected accuracy of the approximation. When the variance parameters $\Sigma$ and $\phi$ are not well-estimated, they are less likely to be consistent across CV folds; in these cases, we expect a large standard deviation among the AXE $\text{LRR}_{j}$'s across CV folds due to the instability of the variance parameter posterior means. In practice, we recommend selecting those clusters with particularly large $n_j$, or for whose estimated random effects are more extreme, and if the mean or standard deviation of $\text{LRR}_{j}$'s is over some threshold $\delta$, we recommend running MCV. The choice of $\delta$ represents the tolerable degree of error in approximating MCV RMSE. For our examples, we use $\delta = 0.25$ as our threshold value, which translates to a ratio of AXE RMSE to MCV RMSE between 0.78 and 1.28. We refer to this additional validation step in our calculations of time cost as AXE+ in Section \ref{sec:reslts_summary}.

\section{Existing CV approximation methods}
\label{sec:other_lco}
We compare AXE to alternative CV approximation methods, which address these challenges in different ways: ghosting (GHOST) \citep{marshall2003approximate}, integrated importance sampling (iIS)  \citep{li2016approximating, vanhatalo2013gpstuff, vehtari2016bayesian}, and likelihood-based linear approximations \citep{jaeckel1972infinitesimal, giordano2019swiss, rad2020scalable}. 
All LCO methods are summarized in Table~\ref{tab:lco_comparison}. We briefly review those methods in Appendix B.

\begin{table}[!ht]
    \centering
     \caption{Reference table of acronyms used}
    \begin{tabular}{ll}
        \toprule
        Acronym & Method \\
        \midrule
         AXE & Approximate cross-validated mean estimates\\
         GHOST  & Ghosting \\\
         iIS & Integrated importance sampling \\
         Vehtari & Vehtari's method \\
         IJ-C & Infinitesimal jackknife, integrated over held-out $\theta_j$ \\
         IJ-A &Infinitesimal jackknife, integrated over all $\theta$ \\
         NS-C & Newton-Raphson, integrated over held-out $\theta_j$ \\
         NS-A & Newton-Raphson, integrated over all $\theta$ \\
         \bottomrule
    \end{tabular}
    \label{tab: acronyms}
\end{table}

Table \ref{tab:lco_comparison} summarizes the differences among the LCO methods, including distributional assumptions, whether they are biased, and computational complexity. While AXE relies only on the expected values of $\Sigma$ and $\phi$ being similar {between the training data and the full data}, GHOST and iIS methods rely on the densities being the same or similar. Note that GHOST has the strongest assumptions, but if $E(\theta_j | \theta_{-j}, \Sigma) = 0$, the sole assumption it relies on is that $E(\beta_{/\theta} | Y) = E(\beta_{/\theta} | Y_{-\mathpzc{s}_j})$. However, GHOST produces biased estimates, while AXE and iIS do not. 

\begin{table}[!ht]
    \centering
     \caption{{Posterior distribution assumptions} and computational complexity of approximating $E(Y_{\mathpzc{s}_j} | Y_{-\mathpzc{s}_j})$ for each LCO method. Cost of Gibbs sampling for equivalent MCV problem is $\bigo{M(N^3P +  NP^2 + P^3)}$, where $N =$ total number of data points $Y$, $P =$ number of coefficients $\beta$, $P_{\Xi} =$ total number of parameters in the model, $M =$ number of MC samples, $n_j =$ size of test data for $j^{th}$ CV fold. }
    \begin{tabular}{lllll}
        \toprule
        Method & $f(\Sigma,\phi | Y)$ vs $f(\Sigma,\phi | Y_{-\mathpzc{s}_j})$  & $f(\beta_{/\theta} | Y)$ vs $f(\beta_{/\theta} | Y_{-\mathpzc{s}_j})$ & Bias & Time \\
        \midrule
         AXE & $E(\Sigma, \phi | Y) = E(\Sigma, \phi | Y_{-\mathpzc{s}_j})$ & N/A& No & $\bigo{N^2P^2 + JP^3)}$\\
         GHOST  & {$f(\Sigma, \phi | Y) = f(\Sigma, \phi | Y_{-\mathpzc{s}_j})$} & $f(\beta_{/\theta} | Y) = f(\beta_{/\theta} | Y_{-\mathpzc{s}_j})$ &Yes &$\bigo{MJP^3 + N}$ \\
         iIS &{$f(\Sigma, \phi | Y) \approx f(\Sigma, \phi | Y_{-\mathpzc{s}_j})$}&{$f(\beta_{/\theta} | Y) \approx f(\beta_{/\theta} | Y_{-\mathpzc{s}_j})$}&No & $\bigo{MJP^3 + MN}$\\
         Vehtari &{$f(\Sigma, \phi | Y) \approx f(\Sigma, \phi | Y_{-\mathpzc{s}_j})$}& N/A&No &$\bigo{M(N^2P^2+JP^3)}$ \\
         IJ
         & N/A & N/A & No & $\bigo{P_{\Xi}^3 + P_{\Xi}^2N + NP}$\\
         NS
         & N/A  & N/A  & No & $\bigo{JP_{\Xi}^3 + NP}$ \\
         \bottomrule
    \end{tabular}
    \label{tab:lco_comparison}
\end{table}

The derivations of computational complexities are provided in Appendix C. Manual cross-validation (MCV), when using Gibbs sampling of the same problem (if available), is the most expensive with complexity $\bigo{M(N^3P +  NP^2 + P^3)}$, where $M$ is the number of posterior samples. However, many of the examples in \Cref{sec:reslts_summary} were fit using STAN, which uses Hamiltonian Monte-Carlo \citep{girolami2011riemann, carpenter2017stan}, where instead of the proposal distribution being a Gaussian random walk, proposal samples are generated along the gradient of the joint density. This allows for more efficient sampling and much shorter run times. In our examples, Vehtari was often the method that took the longest to run, not MCV. iIS and GHOST can also be computationally expensive, typically due to the inversion of $\Sigma$ that is necessary to obtain $E(\theta_j | \theta_{-j}, \Sigma)$ which contributes the $\bigo{P^3}$ term. This cost can be greatly reduced if $E(\theta_j | \theta_{-j}, \Sigma) = 0$ and $\Sigma = \sigma^2I$ for some hyperparameter $\sigma$, $\bigo{MNJ}$, which is the case when the $\theta_j$, $j = 1, \dots, J$, are independent and identically distributed.  It brings both methods closer to AXE in terms of computing cost; however, as we show in \Cref{sec:reslts_summary}, AXE is generally both fast and robust. IJ and NS were typically the two fastest methods in our examples; however, the linear approximation can lead to less accurate estimates in smaller data sets.

\section{Results}\label{sec:reslts_summary}

We use publicly available data to compare AXE and the LCO methods described in \Cref{sec:other_lco} {to manual cross-validation (MCV).} Table \ref{tab:data} is a high-level summary of the data sets and models used in our examples. The first three examples are linear mixed models (LMMs); 
the last three are generalized linear mixed models (GLMMs). 
The first four examples include models where $\theta$ are independent and identically distributed (i.i.d.) and $\Sigma$ is a diagonal matrix, while $\theta$ in the last two examples are Gaussian Markov random fields (GMRFs). Three of our examples consist of balanced data sets (Eight schools, SLC, and SRD), while the remainder are imbalanced.\footnote{Code available online as part of supplemental materials.} 

\begin{table}[!ht]
    \centering
   \caption{Summary of data set and model properties. $P=$ dimension of $\beta$, $J=$ number of CV folds and dimension of $\theta$, $N=$ dimension of response vector $Y$, $n_j=$ size of test data in CV fold.}
   \begin{tabular}{llllrrrr}
    \toprule
   Section& Example& Model& $\theta$ & $P$ & $J$ &$N$ & Max $n_j$ \\
    \midrule
  \ref{sec:8sch} & Eight schools & LMM & i.i.d. & 9 & 8 & 8& 1  \\
   \ref{sec:radon} & Radon & LMM & i.i.d.  & 86-88 & 85 & 919 & 116 \\
  \ref{sec:simul} &  Radon subsets & LMM & i.i.d.  & 4-15 & 3 - 12 & 59 - 100 & 23 \\
   \ref{sec:esp} & Esports (ESP) & GLMM & i.i.d. & 197 & 73 & 2160 & 56 \\
  \ref{sec:slc} & Lip cancer (SLC)  & GLMM& GMRF & 58 & 56 &56 & 1 \\
  \ref{sec:srd} &  Respiratory disease (SRD)  & GLMM& GMRF & 1359 & 271 & 1355  & 5 \\
    \bottomrule
    \end{tabular}
    \label{tab:data}

\end{table}

Detailed descriptions of the data and models are in Appendix~D. Here, we briefly describe the first three examples, including multiple transformations of the data and/or multiple models. The Eight schools example includes transformations of the response designed to reduce information borrowing as a data scaling factor, $\alpha$, increases. There are 40 unique $\alpha$, and all results are over the 40 different transformed data sets. The data can be found directly in the original paper \citep{rubin1981estimation}. The Radon example includes three different models applied to the same data set \citep{goodrich2018rstanarm,gelman2007data}.  The Radon subsets example uses the same set of three models and consists of multiple data sets designed to examine the performance of AXE and other LCO methods under a varying number of clusters or percentages of test data. Using the Radon data, we fixed a particular cluster as the test data and randomly selected $J-1$ other clusters as the training data, $J \in \{3, 4, 6, 9, 12\}$, such that the size of test data was some proportion $\delta$ of the full data, $n_j = \delta N$, $\delta \in \{0.3, 0.4, 0.5, 0.6, 0.7\}$. For each combination of $J$ and $\delta$, multiple subsets are selected and evaluated for a total of 1,475 simulated subsets. Results for each example are aggregated over all data sets and models.

The Radon data are available under a GPL($\ge$3) license. ESP is publicly available under a CC-by-SA 3.0 license; the data we used are included in the supplemental material, along with Radon subsets. SLC and SRD are available under a GPL-3 license \citep{lee2018spatio}.

\subsection{LRR}
We compare AXE to the CV approximation methods described in Section~\ref{sec:other_lco} by calculating LRR$_j$ for each method and CV fold $j$, as in (\ref{eqn:lrr}). For IJ and NS, we calculate MCV RMSE using the posterior mode. 

\begin{figure}[!ht]
    \centering
    \includegraphics[width=\linewidth]{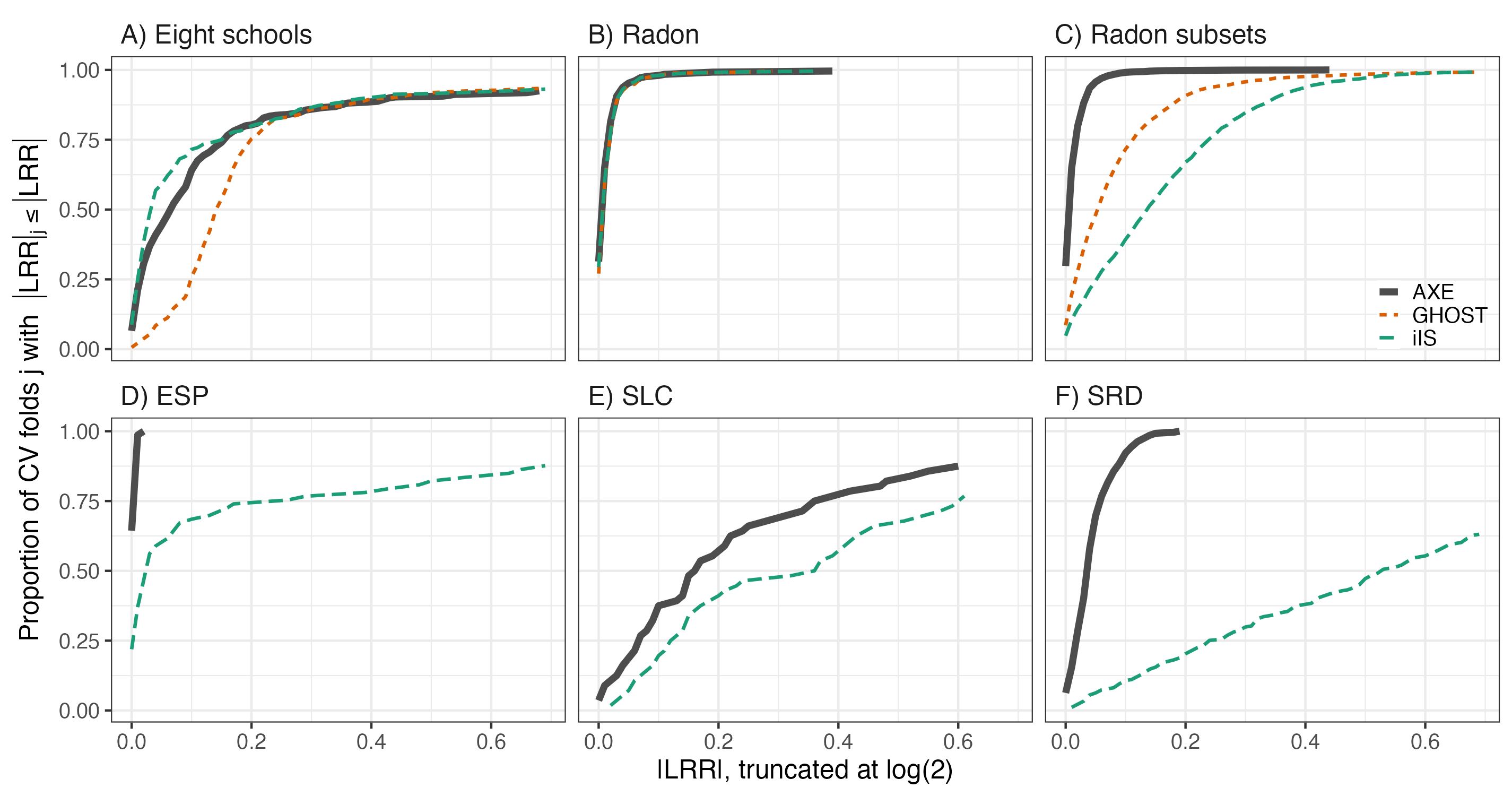}
    \caption{Line plots of the proportion of CV folds $j$ with $|\text{LRR}|_j \le x$, for $x \in [0, \log(2)]$ on the x-axis. We refer to each line as an LRR percentage curve. Curves are colored and shaped based on the method used and are truncated at $\log(2) \approx 0.7$. Ghost is only applicable to LMMs in examples A, B and C.} 
    \label{fig:lrr_perc}
\end{figure}

Figure \ref{fig:lrr_perc} shows the proportion of CV folds with $|\text{LRR}| \le x$ against different cutoff values, $x$. We call these line plots ``LRR percentage curves". A perfectly accurate method would have 100\% of $|\text{LRR}|\text{s} = 0$. An $|\text{LRR}|$ larger than $\log(2)$ corresponds to an approximate RMSE that is over 100\% different from the ground truth MCV RMSE, making it a poor approximation; thus we focus on $x \in [0, \log(2)]$ on the x-axis. Figure \ref{fig:lrr_perc} presents three top-performing methods: AXE, Ghost, and iIS. Comparisons across all methods are provided in Appendix~E. AXE is consistently one of the better-performing methods across our examples, which is not the case for any other method. We examine each panel in detail below.

Figure \ref{fig:lrr_perc}A shows an example where $n_j = 1$ and LCO-CV is the same as LOO-CV. AXE and both iIS methods outperform all others, with roughly 70\% of $|\text{LRR}| \le 0.1$ compared to GHOST's 25\%. As GHOST assumes $f(\beta_{/\theta} | Y) = f(\beta_{/\theta} | Y_{-\mathpzc{s}_j})$, the Ghost mean estimate for this example is $\bar{Y}$, resulting in larger LRRs, while AXE and the iIS methods have mean estimate $\bar{Y}_{-\mathpzc{s}_j}$ (see, e.g., Eight schools in Figure~\ref{fig:axe_each}B).

Figure \ref{fig:lrr_perc}B contains results for the full Radon data, which is a relatively rich data set compared to the number of parameters to estimate. As such, AXE, iIS, and GHOST all have over $97$\% of $|\text{LRR}|$s under $0.1$. A few CV folds do have relatively large $|\text{LRR}|$; these correspond to the same two CV folds across all LCO methods and are cases where the data are particularly informative for the random effect variance. 

Figure \ref{fig:lrr_perc}C contains results for the Radon subsets data, {where the training data are a randomly selected subset of counties.} AXE is the best-performing method with over $99$\% of $|\text{LRR}|\text{s} \le 0.1$. GHOST and iIS perform worse, with $71$\% and $39$\% of $|\text{LRR}|\text{s} \le 0.1$, respectively. Each of those three methods assumes $f(\beta_{/\theta} | Y)$ and $f(\beta_{/\theta} | Y_{-\mathpzc{s}_j})$ are similar, but the low number of {counties} in the training data means that estimation of {the fixed effect coefficient that describes the county-level heterogeneity} can be quite different under cross-validation. We also note that in a closer examination of these results, we found that GHOST tended to underestimate RMSE for the Radon subsets data, while iIS overestimated RMSE. As GHOST uses the full-data posteriors directly, it follows that it produces over-optimistic results. 

In Figure \ref{fig:lrr_perc}D, iIS is able to produce stable estimates because it incorporates information from $\beta_{/\theta}|Y$, but it is outperformed by AXE which has a nearly perfect LRR percentage curve. The good performance of AXE suggests that posterior expectations for $\beta_{/\theta}$ and $\Sigma$ were relatively stable across CV folds.
In panel E, LRRs are in general larger for all methods than preceding panels (see Table~\ref{tab:lrrtable}, in Appendix~F). This is partly due to small sample sizes ($n_j=1$). 
In panel F, AXE yields satisfactory results with $92$\% of $|\text{LRR}|\text{s} \le 0.1$ while the remaining LCO approximation methods perform poorly.

{We summarize each method's performance using the area under the LRR percentage curves (AUC-LRRP) of Figure~\ref{fig:lrr_perc}. An ideal method would have AUC-LRRP of $\log(2)$, and a method with all $|\text{LRR}|$s $> \log(2)$ would have AUC-LRRP of 0. We normalize AUC-LRRP and divide its value by $\log(2)$ so that the maximum is 1 and the minimum is 0. Table~\ref{tab:auc-lrrp} gives AUC-LRRP$/\log(2)$ for each method and example. Further summary metrics for each method and example, consisting of mean LRR and the standard deviation of LRR, are provided in Appendix~F. }

\begin{table}[!ht]
    \centering
        \caption{Area under the LRR percentage curves of Figure~\ref{fig:lrr_perc}, normalized by $\log(2)$ to have a maximum of 1. Methods with highest AUC-LRRP are bolded.}
    \begin{tabular}{lrrrrrrrr}
    \toprule
 & AXE & GHOST & Vehtari & iIS & IJ-C & IJ-A & NS-C & NS-A\\
\midrule
A) Eight schools & 0.80 & 0.72 & 0.16 & \textbf{0.82} & 0.17 & 0.27 & 0.12 & 0.11\\
\hline
B) Radon & \textbf{0.98} & \textbf{0.98} & 0.65 & \textbf{0.98} & 0.70 & 0.88 & 0.71 & 0.47 \\
\hline
C) Radon subsets & \textbf{0.98} & 0.88 & 0.83 & 0.76 & 0.82 & 0.56 & 0.53 & 0.32 \\
\hline
D) ESP & \textbf{1.00} & ** & 0.92 & 0.76 & 0.10 & 0.71 & 0.17 & 0.26 \\
\hline
E) SLC & \textbf{0.67} & ** & 0.35 & 0.52 & ** & ** & 0.19 & 0.34 \\
\hline
F) SRD & \textbf{0.94} & ** & * & 0.33 & ** & ** & 0.12 & 0.09 \\
\bottomrule
\multicolumn{9}{l}{\footnotesize{* Excluded due to computation time.}} \\
\multicolumn{9}{l}{\footnotesize{** Method does not apply.}} \\
    \end{tabular}
    \label{tab:auc-lrrp}
\end{table}

{Table \ref{tab:auc-lrrp} shows that AXE most frequently has the highest AUC-LRRP, except for Eight schools where it had the second highest. The main weakness of Vehtari and iIS is the computational expense. They are the most expensive of all the methods and, in some of our examples, require more time than manual cross-validation, as discussed in Section \ref{sec:results_computation}.}

Both NS and IJ performed worse than the majority of other CV approximation methods. We note that our examples have fewer observations in comparison to the number of model parameters than is typical for NS or IJ, which is usually applied to data large enough such that cross-validation using standard optimization methods is prohibitively expensive. In such situations, the large amount of data typically means that posterior densities are very narrow and $f(\beta_{/\theta}, \Sigma, \phi| Y)$ is more similar to $f(\beta_{/\theta}, \Sigma, \phi | Y_{-\mathpzc{s}_j})$. When there are fewer observations and, subsequently, more uncertainty in the model, a small change in $w$ can result in a larger change in the log-likelihood. Both methods are often limited to LOO-CV, as in \citet{wang2018approximate, rad2020scalable, giordano2019swiss} and \citet{stephenson2020approximate}, which typically results in smaller changes to the log-likelihood over CV folds. We note that all of the CV approximation methods included in this paper improve as the amount of data increases; the main advantage for IJ and NS is that their computing times may scale more easily to LOO-CV with large data sets.

\subsection{Point-by-point comparisons}

\begin{figure}[!t]
    \centering
    \includegraphics[width=\linewidth]{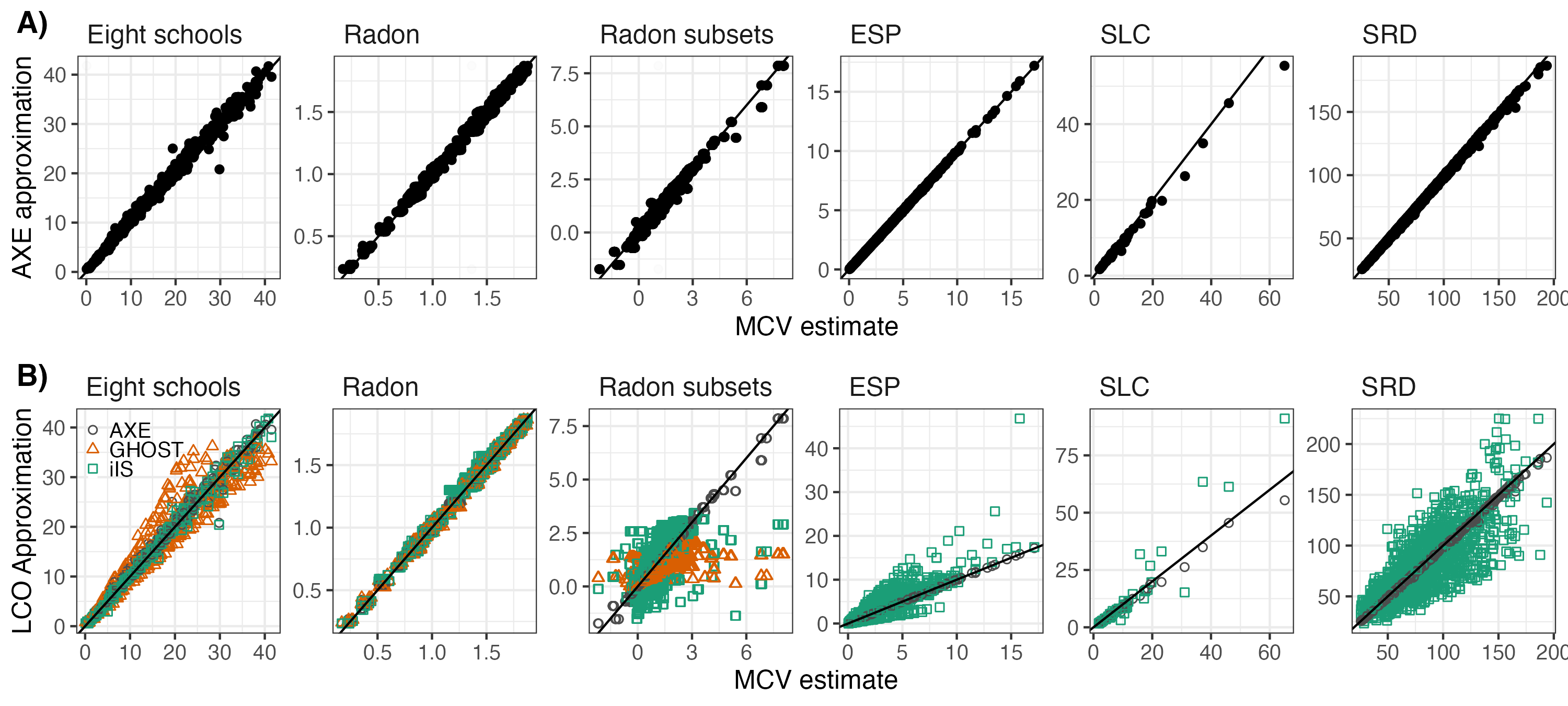}
    \caption{Scatter plots comparing the LCO approximation to ground-truth MCV estimate for each data point, model, and data set. Panels in row A compare the AXE approximation $\hat{Y}_{ji}^{\text{AXE}}$ against the MCV estimate for $E(Y_{ji} | Y_{-\mathpzc{s}_j})$. Panels in row B add points with GHOST (pink triangle) and iIS (green square) approximations whenever applicable, along with AXE (black circle). Each point in a grid represents one point in the corresponding data set(s) and model(s).}
    \label{fig:axe_each}
\end{figure}

We provide graphical point-by-point comparisons of $\hat{Y}^{(\text{method})}$ to ground-truth cross-validated $\hat{Y}^{(\text{MCV})}$. Figure \ref{fig:axe_each}A contains scatter plots of the AXE approximation for each data point $\hat{Y}$ against actual MCV values for all examples. The vast majority of points lie on or near the 45-degree line. In many cases, the AXE approximation is point-by-point equivalent to MCV estimates. It is in contrast to the scatter plots in Figure \ref{fig:axe_each}B, which contain AXE approximations as well as GHOST and iIS, both of which performed relatively well, based on Figure \ref{fig:lrr_perc}. iIS has a higher variance than AXE for the Radon subsets, ESP, and SRD data sets, with approximations farther from the 45-degree line. GHOST has a higher variance in the Eight schools and Radon subsets data sets. All three methods perform similarly in the Radon data sets. 

We found that the accuracy of AXE was not particularly impacted as the proportion of test data increased, but it improved greatly as the number of clusters increased. For instance, some of the largest deviations from the 45-degree line are in row B, the Radon subsets data, where the number of clusters ranges from 3 to 12. It is intuitive, as the estimation of $\Sigma$ depends largely on cluster-specific intercepts, so when there are fewer clusters, $E(\Sigma | Y_{-\mathpzc{s}_j})$ is less stable across CV folds.

\subsection{Computation time} \label{sec:results_computation}
Table \ref{tab:time_comparison} provides the total computation time on an Intel i7-8700k CPU with 40 Gb of memory, {to aid with running examples in parallel; note a standard 8 Gb of memory would be sufficient for the examples.} Except for the SRD data, models were fit using \texttt{STAN} or the \texttt{rstanarm} package, with four chains of 2000 samples and a 1000-sample burn-in, resulting in 4000 total samples. For the SRD data, the \texttt{CAR.ar()} function from the \texttt{CARBayesST} package was used, for four chains of 220,000 samples each, with a 20,000 sample burn-in and thinned to produce 4000 total posterior samples. Calculations for IJ were conducted in Python using the \texttt{autograd} package. AXE took a reasonable amount of time for all examples: a few seconds for Eight schools, Radon, Radon subsets, and SLC, about one minute for ESP, and 10 minutes for SRD.
To date, Vehtari has been used only in the LOO-CV case \citep{vehtari2016bayesian, vanhatalo2013gpstuff}. When CV folds contained multiple data points, Vehtari took much longer than other approximation methods.

\begin{table}[!ht]
    \centering
  \caption{Total computing time for each method in seconds, excluding time to fit the full data. Times with ``h" are in hours. Times to fit the model to the full data are included for comparison.}
\begin{tabular}{lrrrrrrrrr}
\hline
 & AXE & GHOST & Vehtari & iIS & IJ-C & IJ-A & NS-C & NS-A & MCV\\
\hline
A) Eight schools & 0.7 & 1.5 & 1.3h & 0.6h & 0.8 & 0.3 & 4.1 & 1.2 & 0.3h\\
\hline
B) Radon & 9.1 & 0.5 & 14.5h & 476.8 & 2.0 & 0.1 & 100.5 & 2.1 & 1.1h\\
\hline
C) Radon subsets & 1.3 & 90.4 & 6.2h & 3.6h & 118.1 & 45.7 & 82.7 & 34.3 & 15.7h\\
\hline
D) ESP & 67.7 & ** & 55.9h & 570.0 & 3.1 & 1.5 & 120.5 & 59.2 & 4.4h\\
\hline
E) SLC & 0.1 & ** & 2.8h & 738.2 & ** & ** & 16.6 & 1.4 & 693.8\\
\hline
F) SRD & 624.6 & ** & * & 126.0h & ** & ** & 24.5h & 13.5h & 40.6h\\
\hline
\multicolumn{10}{l}{\footnotesize{*: Method excluded due to computation time. **: Method does not apply.}}
\end{tabular}
\label{tab:time_comparison}
\end{table}

\section{Discussion}
\label{sec:discussion}
AXE is a fast and stable approximation method for obtaining cross-validated mean estimates. AXE is equivalent to a Frequentist cross-validation, but using the posterior means $\hat{\Sigma}$ and $\hat{\phi}$ as plug-in estimates, and benefits from any subsequent computational advances.  
{In addition to linear models, the proposed approximation is applicable to all generalized linear models that use an iterative weighted least square algorithm (IWLS). However, the accuracy of the approximation will also depend on the accuracy of IWLS.}
We show that AXE is more accurate when the number of CV folds is large, which is also when it saves the most time. When variance parameters are not well-estimated, we recommend an approximation diagnosis that runs MCV for a small sample of folds and checks whether the mean and standard deviation of LRRs are low. 

Our empirical results show that AXE consistently performed better than more computationally expensive LCO methods. It is because AXE relies on weaker assumptions than many competing methods, making it more robust. Ghosting estimates often had absolute LRR $>0.25$ when 1) the test data were critical for estimating $\beta_{/\theta}$ and $E(\beta_{/\theta} | Y) \neq E(\beta_{/\theta} | Y_{-\mathpzc{s}_j})$ or 2) the $\theta_j$ were not i.i.d., with $E(\theta_j | \theta_{-j}, \Sigma) \neq 0$. The latter scenario is also when ghosting may be more computationally expensive than MCV if the test data size $n_j > 1$. Importance sampling methods typically required longer computation times than either ghosting or AXE. They performed worse when training data posterior densities differed from the full data posterior densities, resulting in high variance. Although importance sampling has weaker assumptions than ghosting, it may also have higher variance, and there are multiple instances where ghosting outperforms importance sampling in our examples. IJ and NS were fast but consistently performed worse than other methods. In addition, IJ only applies to models where the $Y_i$ are conditionally independent, or $\theta$ are discrete and finite and thus could not be applied to the SLC and SRD examples in this paper. 

Another category of LCO-CV approximation methods for Bayesian models in the literature is called integrated or marginal information criteria. The most notable method is the integrated Watanabe-Akaike Information Criterion \citep{li2016approximating, merkle2019bayesian}. Integrated information criteria are used to approximate the expected log predictive density (ELPD), $\sum_{j=1}^J \log f(Y_{\mathpzc{s}_j} | Y_{-\mathpzc{s}_j})$. They do not produce estimates for $E(Y_{\mathpzc{s}_j} | Y_{-\mathpzc{s}_j})$ and so are omitted from the CV comparisons here.

\section*{Acknowledgments}

This work was supported by the National Institutes of Health (NIH) under grants R56AI120812-01A1, R01AI136664, R01AI170249, and R01HL158963. The authors report there are no competing interests to declare.

\section*{Supplementary Materials}

 \begin{description}
 \item[ReadMe:] File describing contents of supplementary materials (readme.md).
 \item[Appendix:] Appendices containing proof of Theorem 1, descriptions of data used, computational complexity calculations, and other supplemental information referenced in the article (AXE\_appendix.pdf)
 \item[AXE examples:] Contains all data sets and codes to re-create the examples in the article. The directory includes a second readme file which describes all files in the code. (AXEexamples, directory). 
 \end{description}

\bibliographystyle{chicago}
\bibliography{AXErefs}

\end{document}